%% file: GE_FrictionWorm.tex
\documentclass[a4paper, 10 pt, conference]{ieeeconf}
\IEEEoverridecommandlockouts 
\usepackage{amsmath}
\usepackage{amssymb}
\usepackage{cite}
\usepackage{gensymb}
\usepackage{graphicx}
\usepackage{textcomp}
\usepackage{hhline}
\usepackage{epsfig}
\usepackage{epstopdf}
\usepackage{pdfpages}
\usepackage{tikz}
\usetikzlibrary{shapes,arrows,backgrounds}
\usepackage[bookmarks=false]{hyperref}
\hypersetup{
	colorlinks = true,
	linkcolor = black,
    filecolor = magenta,
	urlcolor = cyan,
}

\tikzstyle{block} = [draw, fill=white!100, rectangle, minimum height=1cm, minimum width=1.5cm, line width = 1pt]
\tikzstyle{sum} = [draw, fill=white!100, circle, node distance=1cm, line width = 1pt]
\tikzstyle{input} = [coordinate]
\tikzstyle{output} = [coordinate]
\tikzstyle{pinstyle} = [pin edge={to-,thin,black}]
\tikzset{>=latex}

\graphicspath{{figures/}}
\newcommand{\bs}[1]{\boldsymbol{#1}}  

\newcommand{\tf}[1]{\textrm{#1}}

\renewcommand{\baselinestretch}{0.894} 

\title{\LARGE \bf An Earthworm-Inspired Soft Crawling Robot Controlled by Friction}%
\author{Joey Z. Ge, Ariel A. Calder\'{o}n, and N\'{e}stor O. P\'{e}rez-Arancibia%
\thanks{This work was partially supported by the USC Viterbi School of Engineering through graduate fellowships to J. Z. Ge and A. A. Calder\'{o}n, and a start-up fund to N. O. P\'{e}rez-Arancibia. Additional support was provided by the Chilean National Office of Scientific and Technological Research (CONICYT) through a graduate fellowship to A. A. Calder\'{o}n.}%
\thanks{The authors are with the Department of Aerospace and Mechanical Engineering, University of Southern California (USC), Los Angeles, CA 90089-1453, USA (e-mail: {\tt zaoyuang@usc.edu}; {\tt aacalder@usc.edu}; {\tt perezara@usc.edu}).}}%

\begin{document}

\maketitle
\thispagestyle{empty}
\pagestyle{empty}

\begin{abstract}
	We present the design, fabrication, modeling and feedback control of an earthworm-inspired soft robot that crawls on flat surfaces by actively changing the frictional forces acting on its body. Earthworms are segmented and composed of repeating units called \textit{metameres}. During crawling, muscles enable these metameres to interact with each other in order to generate peristaltic waves and retractable \textit{setae} (bristles) produce variable traction. The proposed robot crawls by replicating these two mechanisms, employing pneumatically-powered soft actuators. Using the notion of controllable subspaces, we show that locomotion would be impossible for this robot in the absence of friction. Also, we present a method to generate feasible control inputs to achieve crawling, perform exhaustive numerical simulations of feedforward-controlled locomotion, and describe the synthesis and implementation of suitable real-time friction-based feedback controllers for crawling. The effectiveness of the proposed approach is demonstrated through analysis, simulations and locomotion experiments.        
\end{abstract}

\section{Introduction}
\label{sec01}
Animal locomotion has long been a source of inspiration for robotic research. In particular, the study of limbless crawling has attracted significant attention during the past few years as the most effective method of traveling on unstructured terrains~\cite{ref01, ref02}. One of the most studied species that travel with a limbless gait is the \textit{nightcrawler}, a type of earthworm (\textit{Lumbricus terrestris}). A typical nightcrawler remains underground during the day and crawls above ground at night. As result of this behavior, they have evolved locomotive mechanisms that enable them to maneuver through their labyrinthine underground burrows and crawl over complex terrains. Specifically, nightcrawlers locomote by employing peristalsis, a motion pattern produced by the coordinated and repeated successive contraction and relaxation of the longitudinal and circular muscles embedded in the animals' \textit{metameres} (independent body segments). This periodic pattern can be thought of as a retrograde wave that travels along an earthworm's body to propel it forward using friction-induced traction~\cite{ref03, ref04, ref05}. In the case of nightcrawlers, traction is modulated employing microscopic bristle-like skin structures called \textit{setae}~\cite{ref06, ref07}.

Versatility, robustness and spatial efficiency make the nightcrawler's peristaltic gait a very attractive natural model for robotic locomotion development. Numerous research projects have focused on creating robots that can replicate these earthworms' peristalsis-based locomotion, adopting a variety of different actuation technologies, including \textit{shape memory alloys} (SMAs)~\cite{ref08, ref09, ref10}, magnetic fluids~\cite{ref11} and electric motors~\cite{ref12, ref13, ref14}. Additionally, recent innovations in fabrication methods have enabled the development of biologically-inspired soft actuators, soft sensors and flexible electronics~\cite{ref15,ref16,ref17}. An earthworm-inspired burrowing robot that incorporates these technologies is presented in~\cite{ref18}. That artificial worm was designed to inspect and clean pipes, so its movements and functionalities are constrained to the interior of tubes with diameters in a limited predefined range. As expected, those prototypes are not capable of crawling on open surfaces, which is the problem addressed by the research presented in this paper.

Here, we introduce a new soft robot capable of crawling on flat surfaces, whose basic conceptual design is inspired by the functionality of the \textit{abstract notion} of a two-metamere earthworm. In this design, in order to produce the peristalsis-based retrograde waves required for crawling, a single central linear pneumatic actuator produces the deformations and forces that emulate the axial actions of metameres during earthworm locomotion. Two \textit{extremal} pneumatic actuators produce and modulate the friction forces necessary to alternately anchor the robot's \textit{extremes} to the ground, which is the crucial action in the generation of friction-based crawling. 

The essential mechanism underlying most forms of terrestrial locomotion is friction. Drawing inspiration from nature, researchers have developed several different methods to exploit friction forces, including gecko-inspired adhesives~\cite{ref19}, microspine-based anchors~\cite{ref20, ref21} and anisotropic friction mechanisms~\cite{ref22, ref23}. In the context of soft robotics,~\cite{ref24} presents a robot that employs materials with different coefficients of friction and a pair of unidirectional clutches to manipulate frictional forces to generate locomotion. In the robot presented here, each extremal actuator, made of silicone rubber, varies the friction coefficient between its surface of contact and the ground by expanding and contracting inside a hard 3D-printed smooth casing. This device is designed such that when the actuator is inflated, its silicone-rubber surface touches the ground, producing high friction. Conversely, when the actuator is deflated, its surface does not touch the ground and only the smooth edges of the casing make contact with the supporting surface, thus producing low friction. 

Friction is a nonlinear phenomenon, and consequently, the complete dynamics of the system presented here is both nonlinear and time-varying. However, by treating the forces generated by the central actuator and the friction forces as inputs, the robot's dynamics can be described by a \textit{linear time-invariant} (LTI) state-space model. This reduced-complexity model enables analysis of the system's controllability and is instrumental in determining that locomotion is not feasible in the absence of friction. We explicitly show that the controllability subspace associated with the zero-friction case contains only states that define a constant position of the system's center of mass with respect to the inertial frame of reference. Further analysis shows that if actuation and friction forces were to be chosen at will, the system would become fully controllable. This finding, despite being based on physically unattainable assumptions, indicates that there exists an infinite number of theoretically feasible gaits, and that biologically-inspired locomotion modes represent only a small set of what is possible to achieve with this framework.

\begin{figure}[t!]
	\vspace{1ex}
	\begin{center}
		\includegraphics[width=0.46\textwidth]{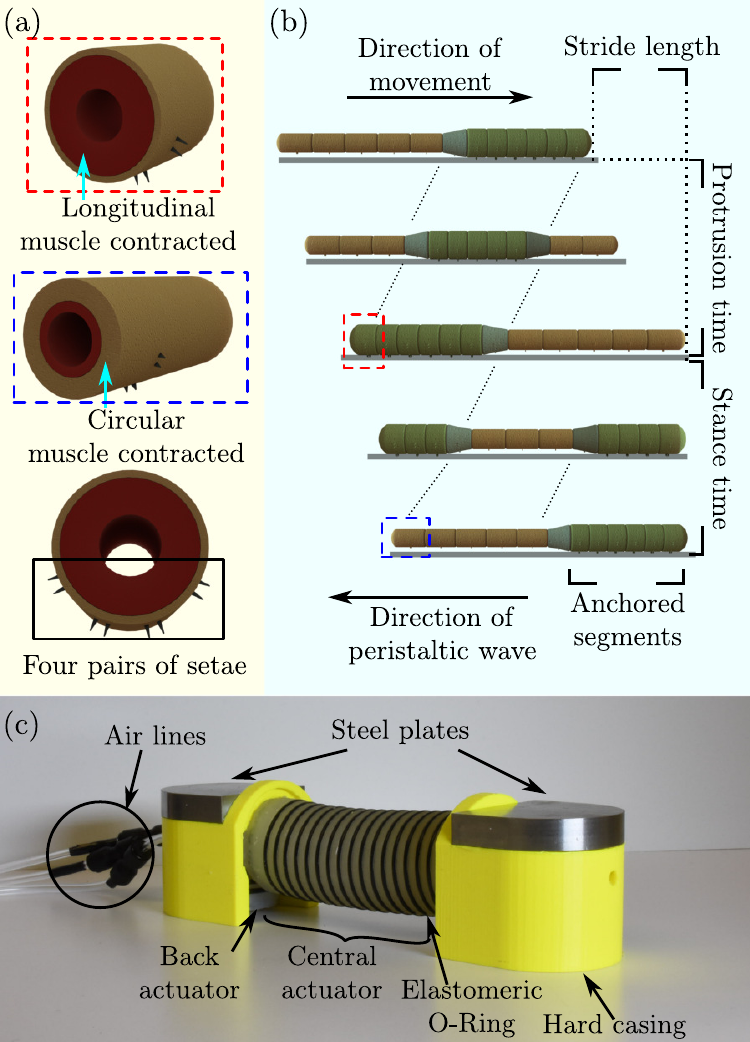}
	\end{center}
	\vspace{-1.5ex}
	\caption{\textbf{(a) Metamere:} A metamere expands radially when its longitudinal muscles contract and expand longitudinally when its circular muscles contract. When a metamere is undergoing radial expansion, the 4 pairs of setae on its ventral and lateral surfaces will protrude and anchor it to the ground. \textbf{(b) Peristaltic crawling motion:} A \textit{stride} is defined as a complete cycle of peristalsis and the \textit{stride length} is the total distance advanced during one stride. The \textit{protrusion time} is defined as the time span during which an earthworm moves forward within a stride. The head of the earthworm covers the stride length by the end of the protrusion time. Correspondingly, the \textit{stance time} is the period during which the head of a earthworm remains anchored to the ground while the rest of the animal body recovers to the initial state. The sum of the protrusion time and stance time is the \textit{stride period}. The thin dotted lines track the retrograde wave. \textbf{(c) Earthworm-inspired crawling robot:} The robot consists of two hard casings, a central actuator, a pair of front and rear actuators constrained by elastomeric o-rings, two machined steel plates, and pneumatic components. \label{fig01}} 
	\vspace{-4ex}
\end{figure}

The rest of the paper is organized as follows. Section~\ref{sec02} introduces the major concepts unique to earthworm-inspired locomotion. Section~\ref{sec03} presents a reduced-complexity dynamic model of the proposed robot and a set of locomotion simulations. Section~\ref{sec04} explains the design and fabrication processes of the soft-robotic components. Section~\ref{sec05} describes the locomotion planning and associated control strategy. Experimental results are presented and discussed in Section~\ref{sec06}. Lastly, Section~\ref{sec07} states the main conclusions of the presented research and provides directions for the future. \vspace{-1ex}
\section{Earthworm-Inspired Locomotion}
\label{sec02}
Earthworms belong to the phylum \textit{annelida}, characterized by their segmented body structures. During locomotion, each ring-shaped segment (metamere) is actively reconfigured by the actions of layers of both longitudinal and circular muscle, as illustrated in Fig.~\ref{fig01}-(a). Internal sealed cavities in earthworms' bodies, referred to as \textit{coeloms}, are filled with incompressible fluid so that the volume of each metamere remains constant while reshaping and the structural integrity of the animal is continually preserved. Anatomical schemes of this type are known as hydrostatic skeletons. Also, the fluid inside each coelom is constrained within each metamere, partitioned by \textit{septa} so that there is no movement of fluid across body segments~\cite{ref05}. Such segmentation preserves, to some extent, the locomotion independence of each metamere, enhancing earthworms' overall mobility~\cite{ref03}. Thus, in order for an earthworm to locomote, the longitudinal and circular muscles of each segment contract alternately, causing each segment to shorten (expanding radially) and elongate (shrinking radially) according to the sequential pattern depicted in Fig.~\ref{fig01}-(b). Such motions from head to tail create the retrograde peristaltic gaits characteristic of worms. It can be proved mathematically that peristalsis-based crawling requires sufficient traction between anchoring metameres and the ground. In the case of \textit{oligochaetas}, the subclass of \textit{annelida} to which earthworms belong, traction is produced and modulated by retractable setae, as depicted in Fig.~\ref{fig01}-(a).  
 
Some species of earthworm are both geophagous (earth-eaters) and surface-feeders~\cite{ref06}. That is the case of nightcrawlers, which emerge from their burrows and crawl on ground at night and remain underground during daytime~\cite{ref25}. To transition and adapt to these two different surroundings, they switch between peristalsis-based crawling and burrowing locomotion modes. A worm-inspired burrowing soft robot was presented in~\cite{ref18} and here we extend that work to the crawling case, which requires the active control of friction. This friction-based control strategy is loosely inspired by the morphology of nightcrawlers, which have evolved setae only on the ventral and lateral surfaces of each metamere to facilitate traction during surface crawling, as illustrated in Fig.~\ref{fig01}-(a). On the other hand, most purely geophagous earthworms have setae arranged in a ring around each body segment~\cite{ref05, ref07}. During crawling, setae protrude from radially expanding metameres (longitudinal muscle contraction) and anchor into the substratum to provide traction, thus preventing slipping while adjacent body segments contract or expand. Once a metamere's circular muscle starts to contract, the longitudinal muscle relaxes and the setae retract from the ground to allow for the segment to slide forward.           

The basic crawling gait of the robot presented in this paper is loosely based on the nightcrawler's crawling mechanism, depicted in Fig.~\ref{fig01}-(b). Following~\cite{ref06}, we define a \textit{stride} as one cycle of peristalsis and describe the crawling kinematics as a function of four variables: \textit{stride length}, \textit{protrusion time}, \textit{stance time} and \textit{stride period} (illustrated in Fig.~\ref{fig01}-(b)). For simplicity, despite the fact that earthworms have numerous segments with staggered stride periods, we define these kinematic variables in relation to an earthworm's first segment. A prototype of the proposed robot is shown in Fig.~\ref{fig01}-(c). This system can be thought of as a two-metamere crawling artificial worm composed of pneumatic soft actuators that emulate earthworms' muscle structures as well as hard casings employed in friction regulation. The processes of locomotion modeling, robotic design, fabrication and controller development are discussed in the next sections.  \vspace{-3ex}
\begin{figure}[t!]
	\vspace{1ex}
	\begin{center}
		\includegraphics[width=0.48\textwidth]{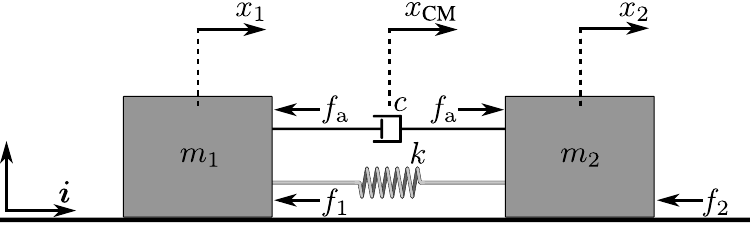}
	\end{center}
	\vspace{-2ex} 
	\caption{Reduced-complexity mass-spring-damper model of the robot in Fig.~\ref{fig01}-(c). A controllability analysis is carried out for two cases: without friction and with frictions $f_1$ and $f_2$ included as inputs. The values of $f_1$ and $f_2$ are positive when the associated vector forces act in the same direction as $\bs{i}$, and are negative, when the vector forces act in the opposite direction as $\bs{i}$. \label{fig02}} 
	\vspace{-3ex}
\end{figure}
\section{Dynamic Modeling and Simulations}
\label{sec03}
\subsection{Robot Dynamics and Controllability Analysis}
\label{sec03a} 
Several of the existing earthworm-inspired robots consist of three sections: a pair of posterior and anterior actuators that serve as artificial circular muscles, and an axial central actuator, which is the analogue of an earthworm's longitudinal muscle~\cite{ref10,ref18,ref26}. Limited by their configurations, those robots can only travel inside pipes with predetermined diameters. Thus, to locomote, a robot of that type replicates the peristaltic burrowing gaits of earthworms according to a scheme in which its anterior and posterior actuators alternately provide anchoring by pressure to the internal surface of a pipe, while its longitudinal actuator extends and contracts to generate displacements along the pipe's axial axis. 

In this section, employing a reduced-complexity dynamic model, linear system theory and experimental data from~\cite{ref18}, we develop the analytical tools necessary to generate a conceptual design for an earthworm-inspired pneumatically-driven soft robot capable of crawling on flat surfaces. An abstraction of this system is shown in Fig.~\ref{fig02}. In this case, given its function and elastic characteristics, a longitudinal actuator is modeled as a massless elastic spring with stiffness constant $k$ and two forces with opposite directions and identical magnitudes $f_\textrm{a}$ (shown in Fig.~\ref{fig02}). The anterior and posterior actuators are modeled as two blocks with masses $m_1$ and $m_2$, capable of varying their friction coefficients with the ground in real time in order to modulate the resulting values of the friction forces $f_1$ and $f_2$. For a pneumatically-driven axial actuator of the type in~\cite{ref18} and Fig.~\ref{fig01}-(c), the magnitude of the produced driving force can be estimated as
\begin{align}
f_{\tf{a}}(t) = s_{\tf{a}} p_{\tf{a}}(t),
\end{align}  
where $p_{\tf{a}}$ and $s_{\tf{a}}$ are the instantaneous internal air pressure and constant cross-sectional area of the actuator, respectively. Note that in this model, in agreement with the experimental data presented in~\cite{ref18}, $k$ is considered to be constant. This approximation is sufficiently accurate for purposes of design, controllability analysis and controller synthesis. However, the true stiffness of the actuator is nonlinear, time-varying and depends on the air pressure inside the soft structure. Lastly, energy dissipation is modeled by a damper with a constant $c$ to be empirically identified.      

To address the problem of controllability, we first consider the frictionless case, in which the force magnitude $f_\textrm{a}$ is the sole input to the system. Thus, by defining $x_1$ and $x_2$ as the position variations of $m_1$ and $m_2$ with respect to an inertial frame of reference, and the corresponding speeds $v_1 = \dot{x}_1$ and $v_2 = \dot{x}_2$, we describe the system with the \textit{single-input--multi-output} (SIMO) state-space realization
\begin{align}
\begin{split}
\dot{x} (t) &= Ax(t)+B_0u_0(t),  \\
y(t) &= Cx (t)+Du_0(t),
\end{split}
\label{eqn:eqlabel{2}}
\end{align}
where 
\begin{align*}
A &=
\left[ 
\begin{array}{cccc}
~0              & ~1              & ~0              & ~0 \\
-\frac{k}{m_1} & -\frac{c}{m_1} & ~\frac{k}{m_1}  & ~\frac{c}{m_1} \\
~0              & ~0              & ~0              & ~1 \\
~\frac{k}{m_2}  & ~\frac{c}{m_2}  & -\frac{k}{m_2} & -\frac{c}{m_2} \\
\end{array}
\right],~
B_0 = \left[\begin{array}{c}   ~0 \\ -\frac{1}{m_1} \\ ~0   \\ ~\frac{1}{m_2} \end{array} \right],\\
C &= 
\left[
\begin{array}{cccc} 
1 & 0 & 0 & 0 \\
0 & 1 & 0 & 0 \\
0 & 0 & 1 & 0 \\
0 & 0 & 0 & 1 \\
\end{array}
\right] ,
~D = \left[\begin{array}{c} 0 \\ 0 \\ 0 \\ 0 \end{array} \right],
~x = y = \left[\begin{array}{c} x_1 \\ \dot{x}_1 \\ x_2 \\ \dot{x}_2 \end{array} \right], 
\label{eqn:eqlabel{3}}
\end{align*}
and $u_0 = f_{\tf{a}}$. In this case, the controllability matrix $\mathcal{C}_{0} = \left[ \begin{array}{cccc} B_0 & AB_0 & A^2B_0 & A^3B_0 \end{array} \right]$ has rank~$2$, thus the system is not controllable, meaning that there exists a set of states that cannot be reached from any possible initial state by the action of input signals~\cite{ref27}. Associated with $\mathcal{C}_{0}$ is the controllable subspace, defined as $\mathcal{C}_{AB_0} = \tf{Image} \left\{ \mathcal{C}_{0} \right\}$, which is equivalent to the set of reachable states from the initial condition $x(0) = 0$, $\mathcal{R}_{\tf{t}}$~\cite{ref28}. It follows that $\mathcal{R}_{\tf{t}} = \mathcal{C}_{AB_0} = \tf{Span} \left\{ \chi_1, \chi_2\right\}$, where
\begin{align}
\chi_1 = \left[ \begin{array}{c}
~1 \\ ~0 \\ -\frac{m_1}{m_2} \\ ~0 \end{array} \right],~
\chi_2 = \left[ \begin{array}{c}
~0 \\ ~1 \\ ~0                \\ -\frac{m_1}{m_2} \end{array} \right].
\end{align}
Therefore, every state in $\mathcal{C}_{AB_0}$ can be written as $\alpha_1\chi_1~+~\alpha_2 \chi_2$ for some $\alpha_1,\alpha_2 \in \mathbb{R}$, which implies that all the reachable positions for the masses take the form $\left\{ x_1 = \alpha_1, x_2 = -\alpha_1 \frac{m_1}{m_2} \right\}$. Thus, we conclude that for all possible inputs and initial state $x(0)=0$, the location of the system's center of mass with respect to the inertial frame remains constant because the variation
\begin{align}
x_{\tf{CM}} = \frac{m_1 x_1 + m_2 x_2}{m_1 + m_2} = 0,
\end{align}     
for all $x \in \mathcal{R}_{\tf{t}}$. The main implication of this analysis is that in the absence of friction, for a robot of the type in Fig.~\ref{fig01}-(c), locomotion is impossible. This result is consistent with generalizable physical intuition and biological observations~\cite{ref29}.    

In the presence of friction, a simple model for the generation of traction forces in the system of Fig.~\ref{fig02} is 
\begin{align}
f_i(t)~=~\tf{sign}\left[\dot{x}_i(t) \right]\mu_i(t)  m_i g,~\textrm{for}~i=1,2, 
\label{EQN05}
\end{align}
where $\mu_i \in \mathbb{R}^+$ are kinetic friction coefficients~\cite{ref30} and $g$ is the acceleration of gravity. Given this structure, the only possible way in which $f_1$ and $f_2$ can be modulated is by varying $\mu_1$ and $\mu_2$. From linear-systems-theory-based analysis it is not possible to determine if the system becomes fully controllable when the inputs are $\left\{f_{\tf{a}},\mu_1,\mu_2 \right\}$. However, we explain the importance of friction for the control of locomotion by analyzing the simplified dynamics resulting from assuming unrestricted inputs $\left\{f_{\tf{a}},f_1,f_2 \right\}$. This case can be modeled by the \textit{multi-input--multi-output} (MIMO) state-space representation $\left\{ A,B_1,C,D\right\}$, where the new input-state matrix and input signal are given by
\begin{align}
B_1 =
\left[ \begin{array}{ccc}
~0 & ~0 & ~0 \\
-\frac{1}{m_1} & -\frac{1}{m_1} & ~0 \\
~0 & ~0 & ~0 \\
~\frac{1}{m_2}  & ~0 & -\frac{1}{m_2}
\end{array} \right],~
u_1 =
\left[ \begin{array}{c}
f_{\tf{a}} \\
f_1 \\
f_2 
\end{array} \right].
\end{align}

For this augmented state-space realization, the associated controllability matrix $\mathcal{C}_1=\left[ \begin{array}{cccc} B_1 & AB_1 & A^2B_1 & A^3B_1 \end{array} \right]$ has rank $4$, and therefore, the controllable subspace $\mathcal{C}_{AB_1}~=~\textrm{Image} \left\{ \mathcal{C}_1 \right\}$ spans $\mathbb{R}^4$. This analysis implies that if the input $u_1$ could be chosen without restriction, any desired state, and consequently, any position of the system's center of mass could be reached in a finite amount of time. In reality, however, $u_1$ is highly restricted by actuator limitations and the nonlinear nature of friction. Despite of these restrictions, controlled locomotion can be achieved by varying the friction coefficients $\mu_1$ and $\mu_2$. This fact is explained using numerical examples in the next section.    
\begin{figure}[t!]
	\begin{center}
		\includegraphics[width=0.48\textwidth]{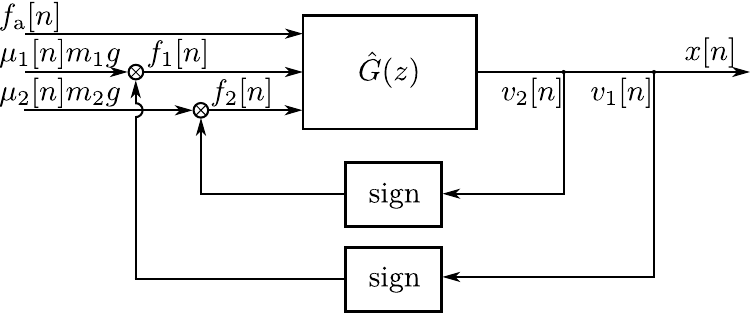}
	\end{center}
	\vspace{-2ex}
	\caption{Discrete-time model used to implement numerical simulations. $\hat{G}(z)$ is the discretized version of $\left\{A, B_1, C, D\right\}$. Sign operators ensure opposing signs between velocities and frictional forces. In this case, both friction coefficients $\mu_i$, $i=1,2$, switch between $\underline{\mu_i} = 0.1$ and $\overline{\mu_i} = 1$. Zero initial conditions are set at the beginning of the simulations. \label{fig03}}
	\vspace{-4ex}
\end{figure}

\subsection{Locomotion Simulations}
\label{sec03b} 
Through numerical simulations, we illustrate how the robot achieves locomotion with the use of feedforward-controlled time-varying friction. Here, a set of feasible control inputs is chosen via an exhaustive search and iteration process. As described in (\ref{EQN05}), the values of the frictional forces $f_i$ are functions of kinematic coefficients of friction $\mu_i$ and the normal forces between the contact surfaces, $m_i g$ (assuming a perfectly flat supporting surface). In the proposed locomotion strategy, normal forces remain constant and friction is regulated by the active variation in real time of the friction coefficients. Specifically, the anterior and posterior actuators of the robot in Fig.~\ref{fig02} are designed and fabricated to switch their coefficients of friction between a small positive value, $\underline{\mu_i}$, and a larger positive value, $\overline{\mu_i}$, in order to produce friction forces with square-wave-signal shapes. This phenomenon is created by actively switching the surfaces of contact between the actuators and supporting ground. The magnitudes of friction coefficients depend on the materials of the surfaces in contact and range from $\sim \hspace{-0.4ex} 0.04$ for Teflon on steel to $\sim \hspace{-0.4ex} 0.8$ for rubber on concrete~\cite{ref31}. According to experimental tests performed on the extremal friction-varying actuators of the robot in Fig.~\ref{fig01}-(c), the measured transition between $\underline{\mu_i}$ and $\overline{\mu_i}$ can be as fast as $0.4~\textrm{s}$, which enables the design and implementation of control strategies based on low-frequency \textit{pulse width modulation} (PWM). For the simulations, we assume that these transitions are instantaneous. 

\begin{figure}[t!] 
	\centering
	\includegraphics[width=0.44\textwidth]{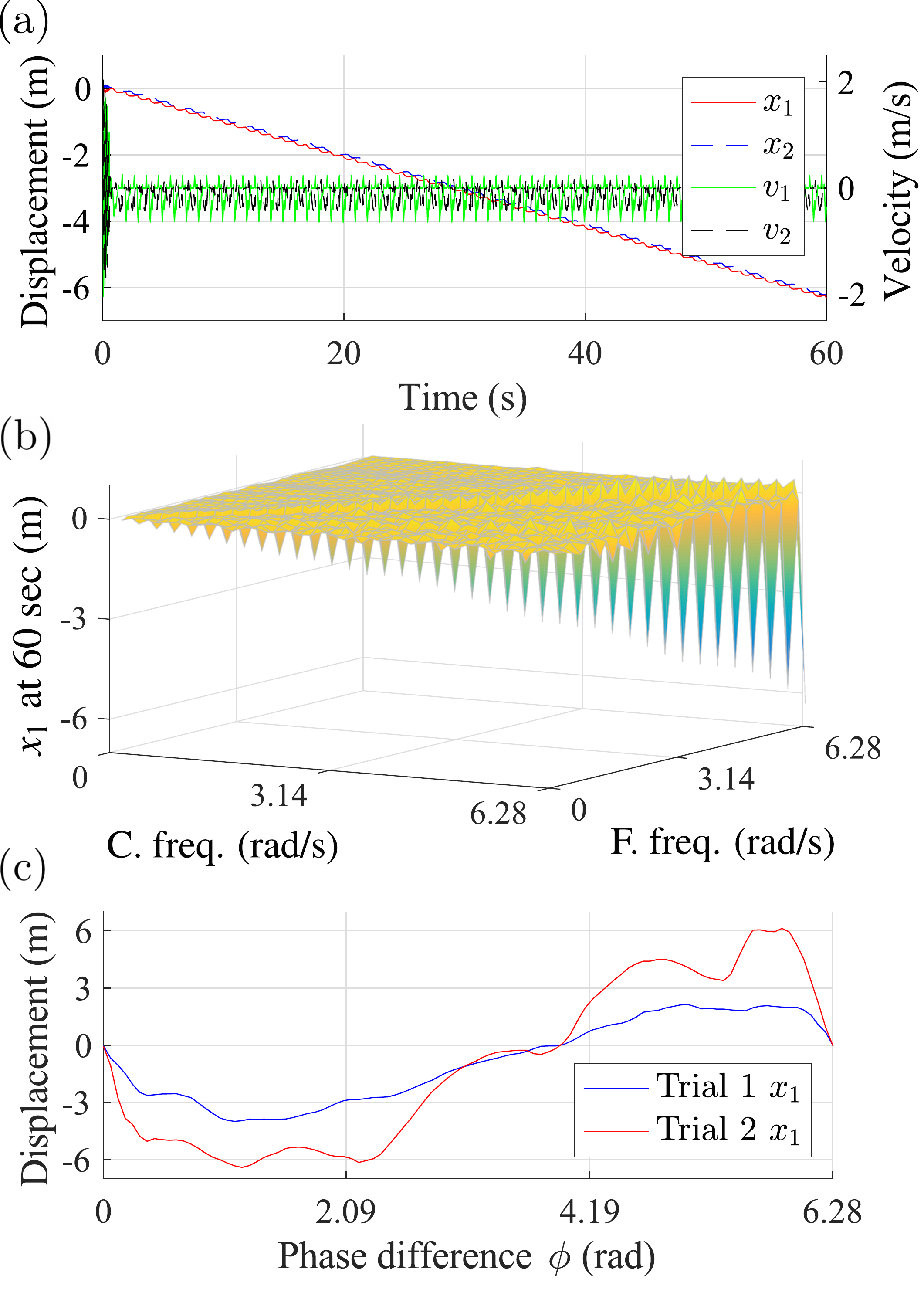}	
	\vspace{-1.5ex}
	\caption{\textbf{(a)} Simulated displacements and instantaneous velocities of $m_1$ and $m_2$ when the frequencies of $f_\textrm{a}$, $f_1$ and $f_2$ are set to 1 Hz and $\phi = 0.4\pi~\textrm{rad}$ ($m_1 = m_2 = 0.2~\textrm{Kg},~k = 200~\textrm{N}\cdot \textrm{m}^{-1},~c=0$). \textbf{(b)} Simulated displacement of $m_1$ at 60 s across a variety of frequency combinations for $f_\textrm{a}$ (C. freq.) and $f_1$ (F. freq.). The forces $f_1$ and $f_2$ oscillate at the same frequency and $\phi$ is held at $0.4\pi~\textrm{rad}$ ($m_1 = m_2 = 0.2~\textrm{Kg},~k = 200~\textrm{N}\cdot \textrm{m}^{-1},~c=0$). \textbf{(c)} Relationship between displacement and phase difference $\phi$ when $f_\textrm{a}$, $f_1$ and $f_2$ are synchronized (at 1 Hz). It can be observed that the direction of locomotion can be reversed by controlling the phase difference between $f_\textrm{a}$ and $f_1$, $f_2$. Trial 1 and Trial 2 correspond to mass values of 0.1~Kg and 0.2~Kg, respectively ($k = 200~\textrm{N}\cdot \textrm{m}^{-1},~c=0$). Heavier masses correspond to higher frictions. These simulation results suggest that higher friction produces faster locomotion. \label{fig04}} 
	\vspace{-3.5ex}	
\end{figure} 

Thus, by combining the actuation model for the generation and control of friction forces with the system dynamics discussed in Subsection~\ref{sec03a}, we implement numerical simulations aimed to study the dynamic behavior of the soft robot during surface crawling. This study is relevant for the search of feasible and, eventually, optimal locomotion patterns. The basic simulation scheme is shown in Fig.~\ref{fig03}, where $\hat{G}(z)$ is the discretized version of $\hat{G}(s) = C \left(sI - A \right)^{-1}B_1 + D$, with the state-space representation $\left\{A_{\tf{D}},{B_1}_{\tf{D}},C_{\tf{D}},D_{\tf{D}} \right\}$, obtained with the \textit{zero-order hold} (ZOH) method and employing a sampling frequency of $1~\hspace{-1.6ex}~\tf{KHz}$. Consistently, the sequences $f_{\tf{a}}[n]$, $f_1[n]$, $f_2[n]$, $x_1[n]$, $x_2[n]$, $v_1[n]$ and $v_2[n]$ are the discrete-time versions of the functions $f_{\tf{a}}(t)$, $f_1(t)$, $f_2(t)$, $x_1(t)$, $x_2(t)$, $v_1(t)$ and $v_2(t)$. 

\begin{figure*}[t!]
	\vspace{-1ex}
	\centering
	\includegraphics[width=1\textwidth]{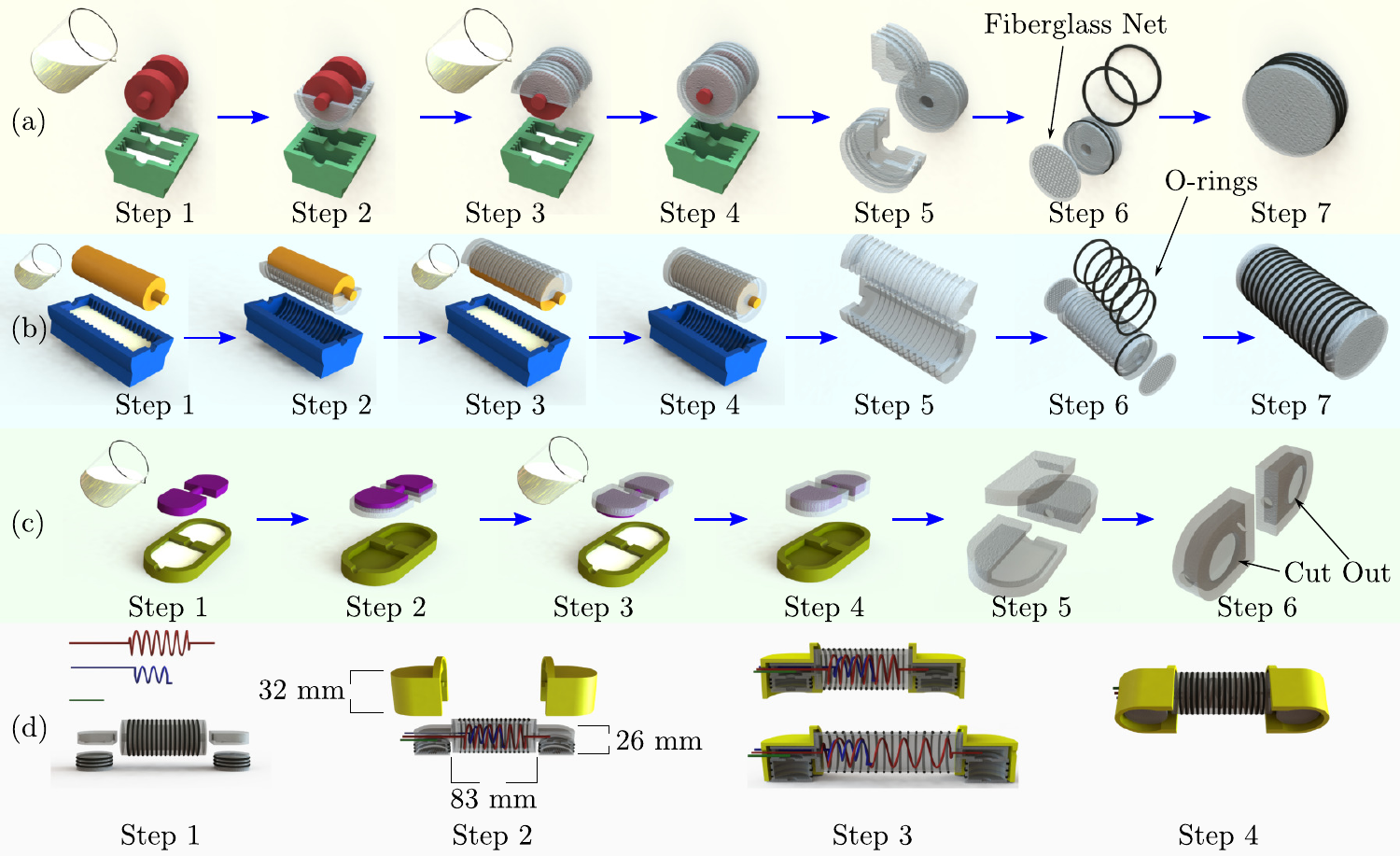}
	\vspace{-2.5ex}
	\caption{\textbf{(a) Fabrication of the front and rear actuators:} First, liquid silicone (Ecoflex\textsuperscript{\textregistered} 00-50, Smooth-On) is poured into a 3D-printed mold and the lower half of a symmetrical double-cylindrical core is submerged in the silicone (Step 1). The silicone within the mold is then cured at $65^{\circ}$ for 15 minutes. The cured silicone half-shell, together with the core attached to it, is extracted from the mold (Step 2). Afterwards, the core is rotated $180^{\circ}$ and liquid silicone is added into the mold again to cast the other half of the shell (Steps 3 and 4). The completed shell is then peeled off the core and Step 5 illustrates the shell's structure. In Step 6, butadiene o-rings are fitted onto the shell's imprinted grooves and a layer composed of silicone and a fiberglass net is applied to seal off one end of the shell. The functions of the reinforcement layer and the o-rings are thoroughly explained in~\cite{ref18}. Step 7 shows one completed front/rear actuator. \textbf{(b), (c) Fabrication of the central actuator and the connecting modules: } These procedures are identical to those employed in the fabrication of the front and rear actuators with the exception that the connecting modules do not require an additional reinforcement layer nor o-rings. In addition, a hole is cut out from the bottom of both connecting modules to allow air flow into both the front and rear actuators in the final assembly. \textbf{(d) Final assembly:} In Step 1, two connecting modules, a pair of extremal actuators, a central actuator and three air feeding lines (two of them have helical shape) are integrated together. A pair of casings are then fixed over the extremal actuator modules (Step 2). Step 3 depicts the robot in two states: when all of its actuators are either uninflated or inflated. Note that the helical structure of the two air-feeding lines makes possible their joint simultaneous contraction or expansion with the central actuator. All these components are glued to each other and sealed by applying and curing extra liquid silicone between all the interfaces. \label{fig05}}
	\vspace{-3.5ex}
\end{figure*}

Assuming a periodic oscillation of the robot's axial actuator, a sinusoidal signal, with amplitude and bias determined by the actuator's minimum and maximum internal pressures, is chosen as input $f_{\tf{a}}$. The extremum air-pressure values are estimated from the experimental data published in~\cite{ref18}. The signals $f_1$ and $f_2$ are chosen to have square-wave shapes with amplitudes and biases given by the lower and upper bounds of the frictional forces associated with the lowest and highest friction coefficients, $\underline{\mu_i}$ and  $\overline{\mu_i}$, respectively. For consistency with the experimental behaviors of the robot's pneumatic actuators, the simulation inputs are limited to a frequency of $1~\tf{Hz}$. Also, $f_1$ and $f_2$ are set to have the same frequency but set apart with a phase difference $\phi$ that can be varied between $0$ and $2\pi~\tf{rad}$. Additionally, because kinetic friction always opposes an actuator's motion, two sign operators are inserted in a feedback configuration, introduced to ensure opposing signs between velocities and frictional forces, as shown in Fig.~\ref{fig03}.  

A set of simulation results is presented in Fig.~\ref{fig04}. Here, for all the cases, we set $k = 200~\tf{N} \cdot \tf{m}^{-1}$, $c = 0 $, $\underline{\mu_1} = \underline{\mu_2} = 0.1$ and $\overline{\mu_1} = \overline{\mu_2} = 1$. Fig.~\ref{fig04}-(a) shows the displacements and velocities of the two mass-blocks ($m_1 = m_2 = 0.2~\tf{Kg}$), when $f_{\tf{a}}$, $f_1$ and $f_2$ oscillate at $1~\tf{Hz}$ and $\phi = 0.4\pi~\tf{rad}$. Both masses travel in an approximately linear motion at an average speed of $6.31~\tf{m} \cdot \tf{min}^{-1}$ ($10.52~\tf{cm} \cdot \tf{s}^{-1}$). Fig.~\ref{fig04}-(b) shows the total distance traveled by the robot in $60~\tf{s}$ versus the frequencies of $f_\textrm{a}$ and $f_1$, $f_2$. All the simulations in this plot were run with a constant phase difference $\phi=0.4\pi~\tf{rad}$ and $m_1 = m_2 = 0.2~\tf{Kg}$. These simulation results suggest that for this specific selection of inputs, substantial locomotion can only be attained when the input frequencies are equal, with exception of a few frequency combinations. Also, in general, faster inputs generate faster locomotion. Fig.~\ref{fig04}-(c) shows the final position reached by the robot after $60~\tf{s}$ across all $\phi$, when all the input frequencies are held at $1~\tf{Hz}$, for two different choices of the pair $\left\{m_1,m_2 \right\}$. This plot suggests that, for this particular type of inputs, $\phi$ is critical for locomotion generation and direction reversal can be realized simply by varying the phase difference between friction inputs. Similar results have been observed in repeated simulations for friction forces with different amplitudes. These findings are limited to the specific set of inputs employed in the discussed cases, but nonetheless exemplify the challenges and potential of friction-controlled crawling. A more comprehensive study of input signals and control strategies to optimize locomotion is a matter of current and future research.

\begin{figure*}[t!]
	\vspace{-1ex}
	\begin{center}
		\includegraphics[width=1\textwidth]{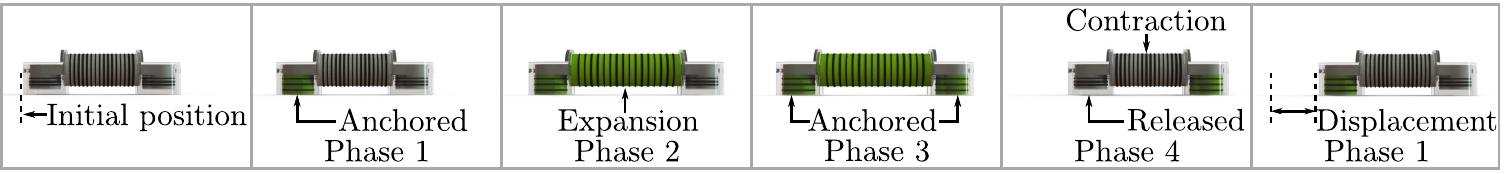}
	\end{center}
	\vspace{-2ex}
	\caption{Actuation sequence for locomotion. Green represents inflation and gray represents deflation. When none of the actuators is inflated, both the front and rear casings are in contact with the supporting surface. In Phase 1, the rear actuator is inflated to make contact with the surface and anchor itself. In Phase 2, the central actuator expands, driving the front actuator forward while the rear actuator remains anchored to the ground. Once the central actuator stops expanding, the front actuator inflates (Phase 3) to anchor itself to the surface. In Phase 4, both the rear and central actuators contract. Such a sequence defines one cycle of the robot locomotion pattern. \label{fig06}} 
	\vspace{-3ex}
\end{figure*}

\section{Design and Fabrication}
\label{sec04}
The work presented in this paper extends that of~\cite{ref18}. The earthworm-inspired soft robot introduced therein can only function constrained by the specific geometric configuration of pipes, employing a burrowing gait. Here, we develop the design, fabrication and control tools necessary to create an earthworm-inspired soft robot capable of crawling on flat surfaces. The key design innovation introduced in this work is the switching of friction forces by alternating the actuators' surfaces of contact with the ground. To achieve such objective, we design the soft robot shown in Fig.~\ref{fig01}-(c), composed of a central longitudinal actuator, a pair of extremal longitudinal actuators, and a pair of hard casings that enclose the extremal actuators. In addition, a pair of soft modules, shown in Fig.~\ref{fig05}-(c), are employed to connect the central actuator with the extremal actuators. These two connecting modules are  also enclosed within the hard casings. 

In the proposed robotic design, actuators are driven pneumatically. The central and extremal actuators are designed to emulate the earthworm's longitudinal and circular muscles, respectively. All actuators are built to expand and contract axially as functions of their internal pressures, unlike those in~\cite{ref18}. Both front and rear actuators are fixed to the upper interior surface of the hard casings and remain above the ground when deflated as the hard casings support the robot's weight. When inflated, the front and rear actuators elongate and make contact with the surface. The hard casings provide low friction while the actuators yield high friction with the supporting surface. Thus, in this scheme, switching between high and low frictional force values is made possible by a simple inflation and deflation sequence. This actuation method is inspired by the traction variable mechanism employed by nightcrawlers, discussed in Section~\ref{sec02}. To see this, recall that, when crawling, their contracted longitudinal muscles (coupled with relaxed circular muscles) will cause a metamere to expand radially, pushing the setae into the ground to anchor and prevent backward slippage. Note that, even though the extremal actuators together with their casings are inspired by earthworm's circular muscles and setae, the underlying working principles are significantly different. In addition, deformation of natural muscles is achieved through active contraction and passive elongation as opposed to the artificial actuators discussed here that elongate actively but contract passively. 
\begin{figure}[t!] 
	\vspace{1ex}
	\begin{center}
		\input{figures/fig07}
	\end{center}
	\vspace{-2ex}
	\caption{LTI scheme used to independently control all the actuators ($j=1,2,3$) of the soft robot shown in Fig.~\ref{fig01}-(c). $p_{r,j}$ is the reference pressure. $\hat{K}_j$ is a PID controller tuned for each actuator. $P_j$ represents each valve and pressure sensor, whose input is the duty cycle of an exciting PWM signal and output is the measured pressure $p_{m,j}$. The controllers are tuned so that the \textit{root mean square errors} (RMSEs) remain smaller than $0.02~\tf{psi}$. \label{fig07}}
	\vspace{-3.5ex}
\end{figure}
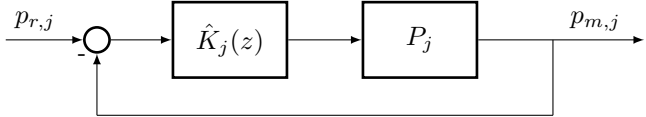 

The methods and construction sequences employed to fabricate the soft robot are depicted in Fig.~\ref{fig05}. Fig.~\ref{fig05}-(a), Fig.~\ref{fig05}-(b) and Fig.~\ref{fig05}-(c) illustrate the fabrication processes of the front and rear actuators, central actuator and the connecting modules, respectively. Fig.~\ref{fig05}-(d) explains the steps leading to the final assembly of the robot. The parts fabricated and materials used to build this robot include 3D-printed \textit{acrylonitrile butadiene styrene} (ABS) molds and casings, silicone elastomer (Ecoflex\textsuperscript{\textregistered} 00-50, Smooth-On), butadiene rubber elastomeric o-rings, fiberglass sheets and pneumatic components. All actuators measure $35~\tf{mm}$ in diameter, the central actuator measures $83~\tf{mm}$ in length and the extremal actuators combined with the connecting modules measure $26~\tf{mm}$ in height. The wall thickness of the soft components range between $2.5$ and $3~\tf{mm}$. These dimensions were chosen based on the robot design in~\cite{ref18}, and were modified to accommodate off-the-shelf pneumatic components. 

To drive the system, an Elemental $\textrm{O}_2$ commercial air pump and a 12-V ROB-10398 vacuum pump are employed to inflate and deflate all actuators through a manifold (SMC VV3Q12). Three high speed solenoid valves (SMC VQ110-6M) and three Honeywell ASDX Series digital serial silicon pressure sensors provide regulation and measurement of each actuator's internal pressure. Data acquisition and signal processing are performed with an AD/DA board (National Instruments PCI-6229) mounted on a target PC which communicates with a host PC via xPC Target 5.5 (P2013b). 

\section{Locomotion Planning and Control}
\label{sec05}

In Section~\ref{sec03b}, using simulations, we demonstrated that fast locomotion is contingent upon perfectly-shaped periodic driving and frictional forces, with perfectly-matched relatively high frequencies. These conditions are not realizable with pneumatically-powered soft actuators as those of the robot in Fig.~\ref{fig01}-(c) (discussed in Section~\ref{sec04}). Thus, replicating the high-speed simulated locomotion behaviors on the actual robot is, at this moment, not an attainable objective. However, we can implement bio-inspired locomotion strategies that are compatible with lower frequencies. It is easy to see from Fig.~\ref{fig02}, that $m_1$ will remain stationary (anchored to the ground) and $m_2$ can slide forward as the central actuator inflates if 
\begin{align}
	|f_1|\geqslant|f_\textrm{a}|>|f_2|
\label{eq07}. 
\end{align} 
The signal $f_1$ corresponds to static friction while $f_2$ is considered to be kinetic friction. Similarly, $m_2$ will be anchored to the ground and $m_1$ will slide forward as the central actuator deflates if 
\begin{align}
	|f_2|\geqslant|f_\textrm{a}|>|f_1|
	\label{eq08}. 
\end{align} 
Here, $f_2$ is a static friction force and $f_1$ is instead, considered to be kinetic friction. Thus, locomotion can be induced by actuating each actuator following a pattern such that the conditions defined in (\ref{eq07}) and (\ref{eq08}) are satisfied in an alternating sequence. In this way, a four-phase actuation sequence is designed to generate one complete stride for the robot as illustrated in Fig.~\ref{fig06}. Before implementing a locomotion sequence, an actuator characterization test is performed to determine a proper set of values for the robot's stride length, stance time and protrusion time. 

\begin{table}[t!]
	\vspace{-1ex}
	\centering
	\renewcommand{\arraystretch}{1.2}
	\caption{Reference pressures during actuation (${psi}$).}
	\vspace{-2ex}
	\begin{tabular}{l*{4}{c}r}
		\textbf{Phase} 	& \textbf{1} & \textbf{2} & \textbf{3} & \textbf{4} \\
		\hline
		Rear Actuator          & 1.2 & 1.2 & 1.2 & 0   \\ 
		Central Actuator       & 0   & 3   & 3   & 0   \\ 
		Front Actuator       & 0   & 0   & 1.2 & 1.2 \\
		\label{tab01}
		\vspace{-2.5ex} 
	\end{tabular}
\end{table}
\begin{figure}[t!]
	\vspace{-2ex}
	\begin{center}
		\includegraphics[width=0.46\textwidth]{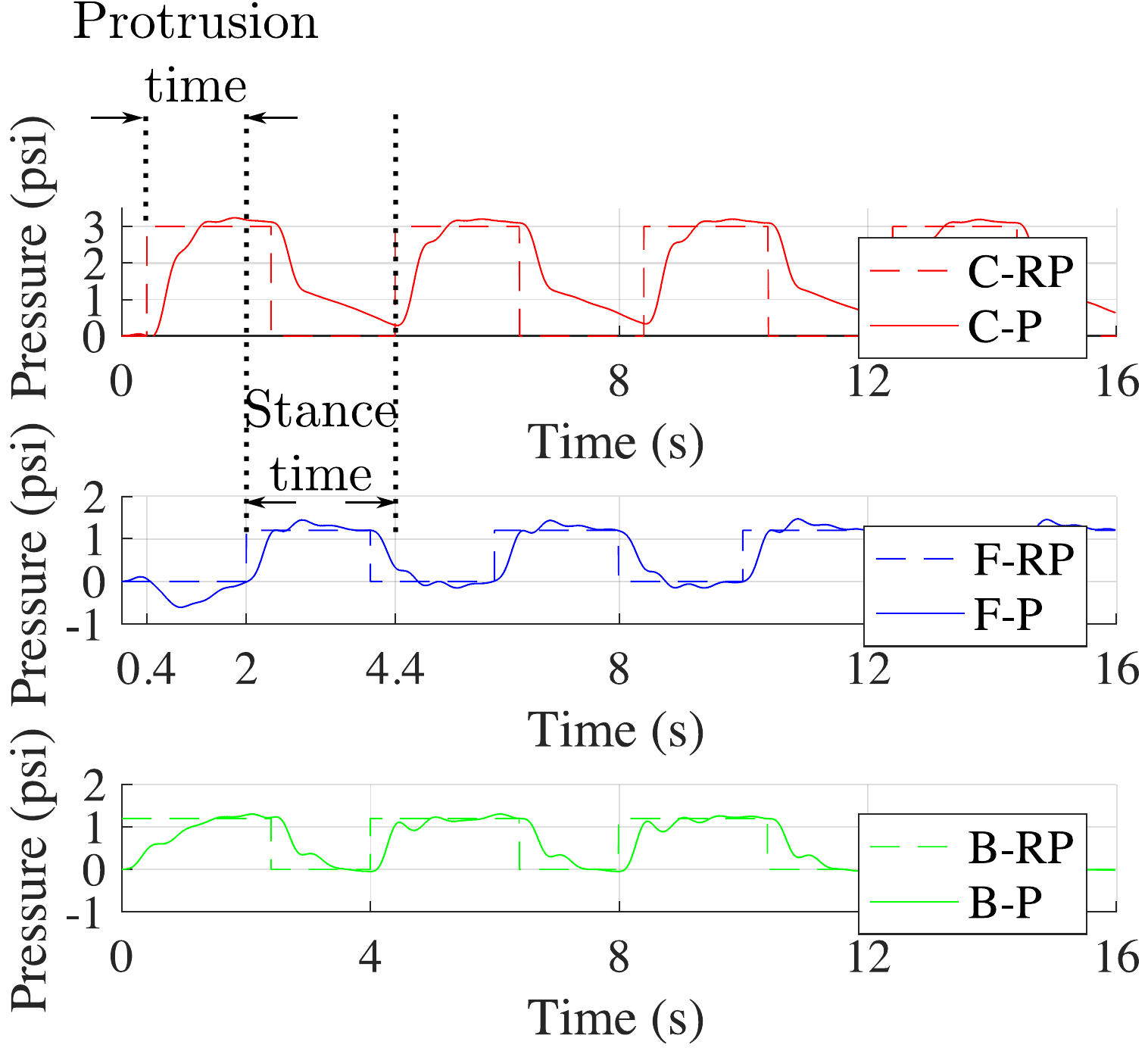}
	\end{center}
	\vspace{-2ex}
	\caption{Example of pressure-tracking experimental results. The continuous lines represent measurements and dashed lines represent references. These data were obtained employing the PID scheme in Fig.~\ref{fig07} to control the central (upper plot), frontal (middle plot) and rear (bottom plot) actuators, during locomotion. The \textit{protrusion time} is $1.6~\tf{s}$, the \textit{stance time} is $2.4~\tf{s}$ and the \textit{stride period} is $4~\tf{s}$. \label{fig08}} 
	\vspace{-4ex}
\end{figure}

To characterize each actuator, three \textit{proportional-integral-derivative} (PID) controllers $\hat{K}_j,~j = 1, 2, 3$, depicted in Fig.~\ref{fig07}, are implemented to regulate internal pressure. Both pumps are maintained at a constant flow rate and output pressure, and the response of each actuator is controlled by solenoid valves using PWM. The valves are normally closed, a state during which the manifold allows for the vacuum pump to deflate the actuators. The PWM duty cycle excites the valves to open and allows for each actuator to inflate individually. In this structure (Fig.~\ref{fig07}), the output of $\hat{K}_j$ is the duty cycle input to each valve. Every PID controller is tuned online in an exhaustive manner. 

The experimental characterization process follows the procedure introduced in~\cite{ref18}. For the central actuator, a range of pressure values that produce substantial elongations without causing significant radial expansions is identified. For the front and rear actuators, the minimum pressure threshold for which firm contact between the actuators and supporting surface is established is chosen to be the reference pressure. Additionally, two 130-gram machined steel plates are fixed onto the top of both casings, as shown in Fig.~\ref{fig01}-(c), to increase frictional force and damp the vibration from the valves during actuation. Table \ref{tab01} presents a set of reference pressures for individual actuators during the four phases described in Fig.~\ref{fig06}. Robot locomotion is achieved by controlling each actuator to track the reference pressure during each phase. In reference to the earthworm crawling kinematics described in Section \ref{sec02}, we define the protrusion time as the period during which the central actuator expands (phase 2). Similarly, the stance time is defined as the time duration after protrusion time during which the front actuator remains static horizontally and completes a cycle of inflation and deflation (phase 3 + phase 4 + phase 1). Protrusion time and stance time are prescribed in experiments. 

\begin{figure}[t!]
	\vspace{-1ex}
	\begin{center}
		\includegraphics[width=0.46\textwidth]{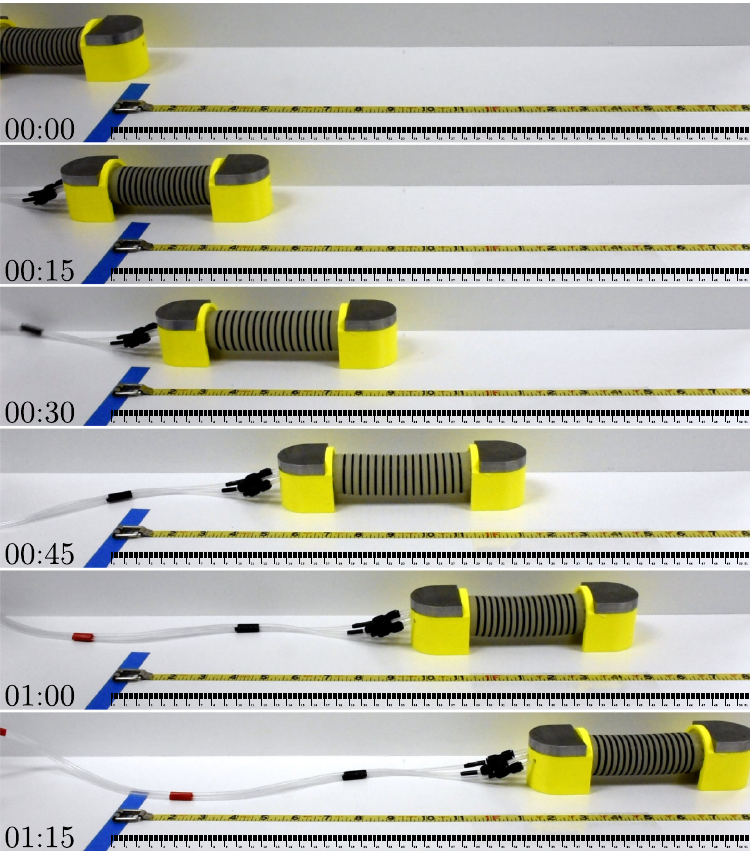}
	\end{center}
	\vspace{-1.5ex}
	\caption{Photographic sequence showing the soft robot while crawling on a laboratory benchtop. Locomotion is achieved by tracking the actuators' pressure references in Table~\ref{tab01}. In this case, a total distance of $52.4~\tf{cm}$ is covered within $75~\tf{s}$ at an average speed of $0.7~\tf{cm}\cdot \tf{s}^{-1}$. The complete set of locomotion experiments can be found in the supporting movie S1.mp4, also available at \href{http://www.uscamsl.com/resources/ROBIO2017/S1.mp4}{http://www.uscamsl.com/resources/ROBIO2017/S1.mp4}. \label{fig09}}
	\vspace{-3ex}
\end{figure}

To implement the described locomotion method, low-level PID controllers (Fig.~\ref{fig07}), tuned during the characterization process, are used to control the actions of each actuator. 

\section{Experimental Results and Discussion}
\label{sec06}
Experiments were conducted to validate the locomotion sequence proposed in Section \ref{sec05}. The first set of tests aims to optimize the crawling speed of the robot on a single uniform surface. The effect of different variables, including the duration of each phase and reference pressures for each actuator, is examined across a broad spectrum of values. Fig.~\ref{fig08} presents the pressure tracking signals of each actuator for the test that produced the fastest locomotion, in which the protrusion time and stance time were $1.6$ and $2.4~\tf{s}$, respectively. A stride length of $2.79~\tf{cm}$ and an average speed of $0.7~\tf{cm} \cdot \tf{s}^{-1}$ were observed and recorded, as shown in Fig.~\ref{fig09}. Since these experiments adopt a different actuation approach to that of the simulations in Section~\ref{sec03}, the large differences between simulated speeds and experimental locomotion speeds are not surprising. As observed in Fig.~\ref{fig08}, the front and rear actuators were able to track the reference pressures with minor overshoots. However, the central actuator was unable to deflate completely. Lower pressure references for the central actuator and longer protrusion times were found to produce better pressure tracking at the cost of overall locomotion speed. No  obvious slippage was observed in any of the tests.

The second set of tests was designed to validate the notion that the robot can travel on surfaces with different coefficients of friction. Using the same actuation sequence than that of Fig.~\ref{fig09}, we proved that the robot can generate peristaltic locomotion on multiple surfaces, including a laboratory benchtop, plywood, \textit{high-density polyethylene} (HDPE), aluminum and a foam pad. Furthermore, we showed that this robot is capable of traversing surfaces with different coefficients of friction by letting it crawl from a foam pad to an HDPE plate. The complete set of all the described tests can be found in the supporting movie S1.mp4, also available at \href{http://www.uscamsl.com/resources/ROBIO2017/S1.mp4}{http://www.uscamsl.com/resources/ROBIO2017/S1.mp4}. 

The experiments presented in this section proved friction manipulation to be an effective way to generate peristaltic crawling in the proposed robot. During locomotion, pressure sensors provide feedback to regulate the elongation of each actuator, and therefore, displacement control was achieved indirectly. Direct displacement control can be implemented in the future by employing a motion-capture system or soft sensors. Also, note that actuator characterization in this case is performed empirically. An analytical model that can capture the nonlinear relationships between an actuator's internal pressure and deformation is needed to improve the control strategy and optimize locomotion. 

\section{Conclusion and Future Work}
\label{sec07}
We presented an earthworm-inspired soft crawling robot capable of locomoting on surfaces by manipulating friction. The robot consists of modular actuators and mechanisms that emulate the functionalities of an earthworm's longitudinal and circular muscles as well as its bristle-like setae structures. We modeled the robot as a mass-spring-damper system and described its crawling dynamics with an LTI state-space representation. We proved mathematically that frictional forces can be employed as inputs that lead to system controllability. This finding was tested and validated through simulations. Experimentally, we demonstrated that the robot is capable of locomoting on surfaces with different coefficients of friction, emulating an earthworm's peristaltic crawling. 

The modular structure of the robot makes it easily scalable, which leaves great potential for creating longer and more versatile robotic structures. Such complex modular systems will provide an ideal platform to develop and test novel decentralized control strategies. In this work, we empirically explored the feasibility of friction-controlled locomotion on flat surfaces. We anticipate that future research will further explore the proposed robotic concept, employing only soft materials and enabling steering and locomotion on uneven terrains. Additionally, the robot presented here is tethered to both the power source and feedback-control module. To achieve autonomy, novel sensing and wireless communication systems must be implemented. Also, portable sources of energy are required. Feasible options are electrolysis and combustion. These topics are a matter of future research.

\bibliographystyle{ieeetran}
\bibliography{REF}

\end{document}

%% file: figures/fig07.tex
	\begin{tikzpicture}
		
		\node[input, name =input1] {};
		\node[sum, right of = input1, node distance = 1.2cm](sum1){};
		\node[block, right of = sum1, node distance = 1.75cm](K){$\hat{K}_j(z)$};
		\node[block, right of = K, node distance = 2.5cm](P){$P_j$}; 
		\node[input, right of = P, node distance = 1.75cm](input2){};
		\node[input, right of = input2, node distance = 1.2cm](input3){};
		\node[input, below of = input2, node distance = 1cm](input4){}; 
		\node[input, below of = sum1, node distance = 1cm](input5){}; 
		
		\draw[->] (input1) node[above right] {$p_{r,j}$} -- (sum1);
		\draw[->] (sum1) -- (K);
	    \draw[->] (K) -- (P);
		\draw[->] (P) -- node[above right] {$p_{m,j}$} (input3){};
	    \draw[-] (input2) -- (input4){};
		\draw[-] (input4) -- (input5){};
		\draw[->] (input5) -- node[above = 0.35cm, left = 0.02cm]{-}(sum1);
		
	\end{tikzpicture}